%% file: main.tex
\documentclass{article}
\usepackage[utf8]{inputenc}
\usepackage[T1]{fontenc}
\usepackage{acl2016}
\usepackage{natbib}
\usepackage{xspace}
\usepackage{graphicx}
\usepackage{amsmath}
\usepackage{url}
\usepackage{lettrine}

\usepackage{natbib}
\setcitestyle{authoryear,open={(},close={)}}

\input{macros}

\aclfinalcopy
\def\aclpaperid{116} 

\title{Bilingual Document Alignment with Latent Semantic Indexing}
\author{%
Ulrich Germann\\
School of Informatics\\
University of Edinburgh\\
\tt{ugermann@inf.ed.ac.uk}}

\begin{document}
\maketitle

\begin{abstract}
We apply cross-lingual Latent Semantic Indexing to the Bilingual
Document Alignment Task at WMT16. Reduced-rank singular value
decomposition of a bilingual term-document matrix derived from known
English/French page pairs in the training data allows us to map
monolingual documents into a joint semantic space. Two variants of
cosine similarity between the vectors that place each document into
the joint semantic space are combined with a measure of string
similarity between corresponding URLs to produce 1:1 alignments of
English/French web pages in a variety of domains. The system achieves
a recall of ca. 88\% if no in-domain data is used for building the
latent semantic model, and 93\% if such data is included. 

Analysing the system's errors on the training data, we argue that
evaluating aligner performance based on exact URL matches
under-estimates their true performance and propose an alternative that
is able to account for duplicates and near-duplicates in the
underlying data.
\end{abstract}

\section{Introduction}
Identifying document pairs that are mutual translations of one another
in large multilingual document collections is an important processing
step in harvesting parallel bilingual data from web crawls. The
\textit{Shared Task on Bilingual Document Alignment} at the
\textit{First Conference on Machine Translation (WMT16)} provides a
common framework to investigate and compare approaches to solving
this problem: given a collection of web site crawls, and a list of
known matches, identify additional document pairs in the collection.

This paper explores the use of cross-lingual {Latent Semantic
  Indexing} \citep{Berry1995} in combination with a URL matching
scheme for this task.

\section{Latent Semantic Indexing}
\subsection{Singular Value Decomposition}
Latent Semantic Indexing \citep[LSI; ][]{dumais88lsa,deerwester90lsa}
is a well-known indexing technique in information retrieval. It relies
on reduced-rank singular value decomposition to map a high-dimensional
\textit{term-document matrix} into a ``semantic'' space of much lower
dimensionality.

The term-document matrix is set up by counting word occurrence in
documents. Each row in the matrix corresponds to a term in the
vocabulary, each column to a document. The individual values in the
matrix are weighted term counts of the respective term in the
respective document. For this work, we use log-normalised term counts
($\mathsf{tf}$ -- {term frequency}) weighted by term
specificity as measured by the inverse document frequency
($\mathsf{idf}$; \citealp{sparck-jones72}\footnote{The technique was
  proposed by Spärck-Jones; the term $\mathsf{idf}$ was coined
  later.}):

\begin{small}
\begin{align}
  w_{t;d} &= \mathsf{tf}\cdot\mathsf{idf}\\
  \mbox{with\quad}
  \mathsf{tf} &= 1 + \log \mathrm{count}_d(t)\\
  \mathsf{idf} &= \log\frac{|\mathcal{C}|}{\mathrm{count}_\mathcal{C}(d: t\in d)}
\end{align}
\end{small}

\noindent
where $t$ is a term from the vocabulary $\mathcal{V}$ and $d$ a document from 
the document collection $\mathcal{C}$.

{Singular value decomposition} \citep[SVD; cf., for example,
][]{manning-schuetze} is then used to factorise this term-document
matrix \M with $m = |\mathcal{V}|$ rows and $n = |\mathcal{C}|$
columns into three matrices $\T_{m\times k}$,$\S_{k\times k}$, and
$\D_{n\times k}$ (with with $k = \min(m,n)$), such that
\T\S\D$^{\T}$=\M.

The column vectors of \T and \D are orthonormal bases of a
$k$-dimensional vector space; \S is a diagonal matrix with the
Eigenvalues of \M in descending order. In other words, dimensions in
which the data differs the most come first, dimensions in which the
data differs little come last. By truncating each of the SVD output
matrices to the respective first $r\ll k$ columns, we obtain a
low-rank representation that approximates the original term-document
Matrix: \mbox{$\T'_{m\times r}\S'_{r\times r}(\D'_{n\times
    r})^{\T}\approx\M$}. (Note, by the way, that $D'\S'\D'^{\T}$ 
is the cosine similarity matrix in the new low-dimensional vector
space.)

\subsection{Document fold-in}

To map a new document into this vector space, we compute the
corresponding new row to be added to $\D'$ as $d_q =
\M_q^{\T}\T'\S'^{-1}$, where $\M_q$ is an additional column in \M that
contains the weighted counts of terms in the respective document.

\section{Alignment of multilingual web pages via cross-lingual LSI}

\subsection{Introduction}
Web pages as delivered by web servers are a mix of data: HTML markup,
which structures the document and pulls in additional resources such
as cascading style sheets, JavaScript libraries, images, and video;
scripts executable in the web browser that influence and extend 
its interactive behaviour and functionality; embedded images and
videos, and, finally, text visible to the human user. Visible text
comprises \textit{boilerplate} and \textit{payload}. Boilerplate text
is text that appears repeatedly across a web site in the form of
menus, page headers and footers, etc. While usually highly
distinctive of a specific web site, boilerplate contributes little to
being able to distinguish individual web pages on a specific site. Web
site readers will usually not pay much attention to boilerplate text
except when navigating the web site; it is nothing that they will
actively read in order to satisfy information needs other than how to
navigate the web site. Payload text, on the other hand, is text that users
visit the specific page for.

While document structure, embedded links, etc. can provide valuable
clues for the alignment of web pages, this work focuses on the text
extracted from the original HTML, as provided by the workshop
organisers as part of the data set. 

\subsection{Approach}
The central idea in our approach is to use cross-lingual LSI to map
monolingual documents into a joint vector space and use similarity
between the corresponding embedding vectors to perform bipartite
alignment of pairs of documents in different languages.


To obtain a cross-lingual model of latent semantics, we first set up a
bilingual term-document matrix \M using parallel documents, keeping
the vocabularies of the two languages separate, so that identical word
forms in the two languages correspond to different rows in \M.
Rank-reduced SVD is then performed on this bilingual matrix to map the
terms of the two languages into a common semantic space with 1,000
dimensions.%
\kern-.25ex\footnote{We used the open-source software package
  \textit{redsvd} \citep[randomised SVD; ][]{Oka10} to perform the
  singular value decomposition.}  Via fold-in, all monolingual
documents from the collection that have been labelled by the language
recogniser as being in one or the other of the language in question
are also be mapped into this common space.

We then use \textit{Competitive Linking} \citep{competitive-linking}
to obtain a bipartite alignment of documents: first, we rank all
possible bipartite alignment hypotheses by score. Processing the list
of hypotheses in descending order, we keep all hypotheses that do not
overlap or conflict with higher-ranking hypotheses and discard the
others. (In fact, competitive linking is what the official evaluation
procedure for this shared task does; for the purpose of participation
in the Shared Task, it is sufficient to produce a ranked list).

\subsection{Term Weighting}

As mentioned above, text extracted from a web page consists of
boilerplate and payload text. To reduce the influence of the former
and boost the impact of the latter on the document vectors, we compute
\idf separately for each domain in the set (rather than globally
across all domains). Thus, terms that occur frequently across a
particular web site will receive a low specificity score (i.e., \idf)
on pages from that web site, yet may receive a high score if they
appear elsewhere.

\subsection{Scoring functions}

In our experiments, we explored and combined the following scoring functions:

\subsubsection{Cosine Similarity (cos)}

This is the classical measure of similarity in LSI-based Information
Retrieval. It computes the cosine of the angle between the two vectors
that embed two candidate documents in the joint semantic vector space.

\subsubsection{``Local'' cosine similarity (lcos)}
The intuition behind the local cosine similarity measure is this:
since we perform SVD on a bilingual term-document matrix that
consists of document column vectors for documents from a large
collection of web sites, web pages from each specific web site will
still appear quite similar if the web site is dedicated to a
particular topic area (which the vast majority of web sites
are). Similarity scores will thus be dominated by the general domain
of the web site rather than the differences between individual pages
within a given web site. The local cosine similarity measure tries
to mediate this phenomenon by shifting the origin of the vector
space to the centre of the sub-space in which the pages of a
particular web site reside before computing cosine similarity. In
practice this is accomplished by subtracting the mean embedding
vector for the domain in question from each individual embedding
vector for pages in that domain. Note that we are only comparing
pages that belong to the same web site within the context of the
shared task.


\subsubsection{URL similarity (url)}
The data provided for the Shared Task contains many duplicates and
near-duplicates of web pages. Duplicates occur when multiple URLs lead
to exactly the same content (e.g. {\tt www.domain.com} and {\tt
  www.domain.com/index.html}); near-duplicates are often the result of
dynamically created content, such as results of database look-up
(e.g., calendars, stock price trackers), embedded page counts, or
different boilerplate due to different language settings delivering
the same payload (e.g., an English article delivered under two
different country-specific user interfaces using different boilerplate
text). Not knowing how the reference set for evaluation within the
Share Task was constructed, we conjectured that the gold standard used for
evaluation might be biased towards URL matches.

Hence, we devised the following match score for pairs of URLs.

\begin{enumerate}
\item All URLs within a domain are tokenised into blocks of either all
  letters or all numbers relying on POSIX UTF-8 character classes;
  punctuation is discarded.
\item For a given pair of candidate URLs, we determine via the
  Needleman-Wunsch algorithm \citep{needleman-wunsch} the cumulative
  score of the longest match sequence between the token sequences
  corresponding to the two URLs. The match score for each individual
  token pair $\tuple{t_1, t_2}$ in the alignment is computed as
  follows. 
  \begin{itemize}
  \item $\mathrm{score}(t_1, t_2) = 0$ if $t_1 \neq t_2$ and at least one of them is a number
  \item $\mathrm{score}(t_1, t_2) = \frac{1}{\mathrm{cnt}(t_1)^2}$ if
    \mbox{$t_1 = t_2$}, where $\mathrm{cnt}(t)$ is the
    position-independent count of token $t$ in all the URLs in the
    collection. The match weighting based on relative frequency in the
    domain serves to discount very frequently occurring URL
    components, (such as \textit{http} or \textit{www}) and boost
    components that are rare in the URLs for this domain, such as, for
    example, article IDs.
  \item
    $\mathrm{score}(t_1, t_2) = \frac{2 * \mathrm{lcss}(t_1,
    t_2)}{\mathrm{len}(t_1) + \mathrm{len}(t_2)} \cdot
    \frac{1}{\mathrm{cnt}(t_1) \cdot \mathrm{cnt}(t_2)}$ if $t_1$ and
    $t_2$ both are sequences of letters, where $\mathrm{lcss}(t_1,
    t_2)$ is the length of the longest common letter sub-sequence
    between $t_1$ and $t_2$. The idea behind this soft match score is
    to reward cognates over candidate pairs that have no semblance of
    one another whatsoever. For example, the lcss score component for
    the pair $\langle \mbox{\itshape London}, \mbox{\itshape
      Londres}\rangle$ would be ca. 0.62 ($\frac{2 *
      4}{\mathrm{len}(\mbox{\tiny\itshape ``London''}) +
      \mathrm{len}(\mbox{\tiny\itshape``Londres''})}$), whereas the
    pair $\langle\mbox{\itshape London}, \mbox{\itshape Paris}\rangle$
    would receive a match score of 0, each of the scores yet to be
    weighted by $\frac{1}{\mathrm{cnt}(t_1) \cdot
      \mathrm{cnt}(t_2)}$. This soft matching score serves to
    accommodate web sites that base their URLs on, for example, the
    headlines of articles or posts.
    
  \end{itemize}
\end{enumerate}

\input{tab-recall}
\section{Evaluation}

\subsection{Recall on training and test data}
To rank alignment hypotheses, we investigated all uniform linear
combinations of the three individual scoring
functions. Table~\ref{tab-recall} shows the results for the training
set, and, in the last row, the performance of the best feature
combination on the test set. In the first set of experiments on the
training set, whose results are shown in the left half of the table,
we used the list of known matches in the training data both for
seeding cross-lingual LSI and evaluation. These numbers give us a
sense how well monolingual documents are mapped into the joint
semantic space by LSI and document fold-in. The first column of the
recall numbers (``strict'') follows the official evaluation procedure,
counting only exact URL matches as correct. The following columns show
the performance if a more lenient notion of ``matching documents'' is
applied. This more lenient measures computes the similarity between
the expected and a proposed target document for a given source
document (and vice versa) as follows:
\begin{align}
  \mathrm{score}(\mathrm{text}_1, \mathrm{text}_2)
  = \frac{2\cdot\mathrm{lcss}(\mathrm{text}_1, \mathrm{text}_2)}%
  {|\mathrm{text}_1| + |\mathrm{text}_2|}
\end{align}

\noindent The length of the longest common sub-sequence (lcss) is here
measured in terms of space-separated tokens as they occur in the
text. No more sophisticated tokenisation is performed. The
content-based evaluation measure counts a proposed match as correct if
the similarity between a proposed target (or source) document and the
expected document is greater or equal to the threshold indicated in
the column header.

The right half of the table shows the results for the same evaluation
performed on the basis of original bilingual term-document matrices
that \textit{exclude} all known matches from the domain in question,
relying only on known matches from other web domains. This leads to
fewer vocabulary matches, as terms specific to the web site in
question may not be included in the model. As expected, we see a drop
in performance, but we are still able to recover about 92.5\% (down
from 96.7\%) of the known matches, even when counting only full
matches and matches with exact duplicates.

\subsection{Error analysis}
\input{tab-misses} Table~\ref{tab-misses} shows the distribution of
missed page pairs over the respective domains in the test data. As we
can see, errors are concentrated in only a few of the 203 domains in
the test set. We will briefly discuss the top five here. The errors in
\texttt{www.lagardere.com} originate from mixed-language pages,
typically pages with the boilerplate text for the user interface
in one language and the actual content in the other. The missed pairs
in \texttt{meatballwiki.org} can be attributed to {\itshape red herrings}:
URL pairs that erroneously suggest a correspondence between the two
pairs in question. The web site \texttt{www.toucherdubois.ca} provides
teaching resources (including images and lesson plans) for teaching
students about ``the sociocultural heritage of the people of
Madawaska'' in Canada and the US. Some of the pages consist of little
text wrapped around image resources; lesson plans are often very
similar in terms of the vocabulary used, thus confusing the LSA model.
The missing pairs from \texttt{www.rfimusique.com} and \texttt{www.taize.fr}
are pairs of pages with a low payload-to-boilerplate (or
near-boilerplate) ratio, i.e., they are dominated by text that can be
found on multiple pages, thus leading to document alignment errors.

\section{Related work}

One of the first systematic approaches to identifying parallel data on
the web is the STRAND algorithm \citep{resnik99strand}. It is a
pipeline process that first generates candidate pairs via a web search
(or by link analysis if a complete download of a web site is
available). It then performs language identification on the retrieved
pages and analyses the HTML structure of candidate documents in order
to filter out document pairs that are too dissimilar in their document
structure. \citet{resnik03web} extend this approach by adding
content-based analysis. They use probabilistic word translation
lexicons to assess the probability that two pages are translations of
each other.

Very similarly to the work presented in this paper, \citet{Saad2014}
use LSI for identification of parallel and comparable corpora. In
addition to the cross-lingual LSI approach taken here, they also
investigate monolingual LSI after document translation. They conclude
that cross-lingual LSI is competitive with monolingual LSI of
automatically translated texts.

\section{Conclusion}

We have investigated the feasibility of using cross-lingual LSI for
identifying parallel documents in large collections of text. Our
results suggest that this is a viable approach to harvesting parallel
data from web crawls. We achieve the best performance with a
combination of classical cosine measure, ``local'' cosine measure, and
URL matching.

The existence of duplicate and near-duplicate documents in the data
raises the question whether it is reasonable to measure performance in
terms of URL matches, or whether evaluation should be based on the
distance between retrieved and expected documents.

\section*{Acknowledgements}
\lettrine[image=true, lines=2, findent=1ex, nindent=0ex, loversize=.15]%
         {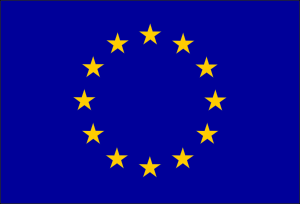}%
         {T}his work was conducted within the scopes of the Innovation Action \textit{MMT} and the Research and Innovation Action \textit{SUMMA}, which have received funding from the European Union’s Horizon 2020 research and innovation programme under grant agreements No 645487 and 688139, respectively.

\bibstyle{author-year}
\bibliography{wmt16}
\end{document}

%% file: macros.tex
\newcommand{\T}{\ensuremath{\mathsf{T}}\xspace}
\newcommand{\D}{\ensuremath{\mathsf{D}}\xspace}
\newcommand{\M}{\ensuremath{\mathsf{M}}\xspace}
\renewcommand{\S}{\ensuremath{\mathsf{S}}\xspace}
\newcommand{\idf}{\ensuremath{\mathsf{idf}}\xspace}

\newcommand{\tuple}[1]{\ensuremath{\langle}#1\ensuremath{\rangle}}

%% file: tab-recall.tex
\begin{table*} 
\caption{Recall on the training and test data with known in-domain document pairs included in / excluded from the initial term-document matrix.}\label{tab-recall}\vspace{1em}

\renewcommand{\tabcolsep}{.4ex}
\begin{tabular*}{\linewidth}{%
*{3}{p{3em}}@{\extracolsep{\fill}}rrrrr@{$\quad$}||rrrrr@{$\quad$}}

\multicolumn{13}{l}{\bfseries performance on the training data}\\\hline
  &&&
  \multicolumn{5}{c||}{\bfseries included}
  &
  \multicolumn{5}{c}{\bfseries excluded}\\
\multicolumn{3}{c}{\textbf{features used}}
& strict\makebox[0pt][l]{\textsuperscript{a}} 
& 1.00\makebox[0pt][l]{\textsuperscript{b}}  
& 0.99\makebox[0pt][l]{\textsuperscript{b}} 
& 0.95\makebox[0pt][l]{\textsuperscript{b}} 
& 0.90\makebox[0pt][l]{\textsuperscript{b}}
& strict\makebox[0pt][l]{\textsuperscript{a}} 
& 1.00\makebox[0pt][l]{\textsuperscript{b}}  
& 0.99\makebox[0pt][l]{\textsuperscript{b}} 
& 0.95\makebox[0pt][l]{\textsuperscript{b}} 
& 0.90\makebox[0pt][l]{\textsuperscript{b}}\\\hline

\multicolumn{3}{c}{cosine (cos)}
&   86.7 & 93.4 & 95.4 & 96.7 & 97.6
&   82.5 & 88.9 & 91.3 & 92.9 & 93.7\\
\multicolumn{3}{c}{``local'' cos. (lcos)}
&   86.7 & 92.8 & 94.7 & 95.8 & 96.9
&   83.3 & 88.9 & 91.4 & 92.8 & 93.6\\
\multicolumn{3}{c}{URL similarity (url)}
&   83.6 & 87.8 & 88.1 & 88.2 & 88.2
&   83.6 & 87.8 & 88.1 & 88.2 & 88.2\\
cos&lcos &
&   87.2 & 93.7 & 95.6 & 96.6 & 97.5
&   83.3 & 89.7 & 92.1 & 93.6 & 94.4\\
cos&&url
&   90.6 & 94.7 & 95.6 & 96.4 & 97.1
&   86.3 & 90.6 & 91.4 & 92.7 & 93.5\\
& lcos & url
&   91.3 & 95.4 & 96.3 & 97.2 & 97.8
&   86.8 & 91.3 & 92.2 & 93.4 & 94.2\\
cos & lcos & url
&   92.8 & 96.7 & 97.6 & 98.5 & 99.1
&   88.0 & 92.5 & 93.4 & 94.7 & 95.5\\
\hline
\multicolumn{8}{l||}{}\\
\multicolumn{8}{l||}{\bfseries performance on the test data}\\
\hline
cos & lcos & url
&&&&&
& 87.6 & 87.6 & 94.1 & 95.5 & 96.0\\
\hline
\end{tabular*}

\begin{tabular*}{\linewidth}{ll}
\textsuperscript{a}& exact string match with the reference ULR pairs\\
\textsuperscript{b}& soft match based on document similarity with different similarity thresholds.
\end{tabular*}

\end{table*}

%% file: tab-misses.tex
\begin{table}[t]
  \caption{Distribution of missed pairs over domains with a soft similarity threshold of .95. Domains with a single miss are aggregated under ``other''.}
  \label{tab-misses}
  \begin{center}
    \begin{tabular}{lr}
      \bfseries domain & \bfseries missed pairs\\\hline
       www.lagardere.com & 20 \\
       meatballwiki.org & 12\\
       www.toucherdubois.ca & 8\\
       www.rfimusique.com & 8\\
       www.taize.fr & 6\\
       www.lalettrediplomatique.fr & 4\\
       www.publictendering.com & 3\\
       www.iisd.ca & 3\\
       hrcouncil.ca & 3 \\
       arabpressnetwork.org & 3\\
       www.technip.com & 2\\
        www.kinnarps.com & 2\\
        www.gameonly.com & 2\\
        www.eufic.org & 2\\
        \itshape other & 17\\
        \hline
    \end{tabular}
  \end{center}
\end{table}